\documentclass[12pt,onecolumn,letterpaper]{article}

\usepackage{cvpr}
\usepackage[utf8]{inputenc}
\usepackage[OT4]{fontenc}
\usepackage{graphicx}
\usepackage{color}
\usepackage[breaklinks]{hyperref}
\hypersetup{
     colorlinks   = true,
     citecolor    = blue
}
\expandafter\def\expandafter\UrlBreaks\expandafter{\UrlBreaks\do\-\do\-}%
\newcommand{\figref}[1]{\hyperref[#1]{\figurename~\ref{#1}}}
\emergencystretch=\maxdimen
\newcommand{\doi}[1]{\href{http://dx.doi.org/#1}{\nolinkurl{doi:#1}}}

\cvprfinalcopy 

\def\cvprPaperID{****} 

\begin{document}

\title{Level generation and style enhancement –- deep learning for game development overview\thanks{Submitted to the 52\textsuperscript{nd} International Simulation and Gaming Association (ISAGA) Conference 2021}}

\author{Piotr Migdał\thanks{\tt{pmigdal@gmail.com}, \url{https://p.migdal.pl}}\\
\small{ECC Games SA}

\and
Bartłomiej Olechno\\
\small{ECC Games SA}

\and
Błażej Podgórski\\
\small{ECC Games SA}\\
\small{Kozminski University}
}

\maketitle

\begin{abstract}
We present practical approaches of using deep learning to create and enhance level maps and textures for video games -- desktop, mobile, and web. We aim to present new possibilities for game developers and level artists.

The task of designing levels and filling them with details is challenging. It is both time-consuming and takes effort to make levels rich, complex, and with a feeling of being natural. Fortunately, recent progress in deep learning provides new tools to accompany level designers and visual artists. Moreover, they offer a way to generate infinite worlds for game replayability and adjust educational games to players’ needs.

We present seven approaches to create level maps, each using statistical methods, machine learning, or deep learning. In particular, we include:

\begin{itemize}
    \item Generative Adversarial Networks for creating new images from existing examples (e.g. ProGAN).
    \item Super-resolution techniques for upscaling images while preserving crisp detail (e.g. ESRGAN).
    \item Neural style transfer for changing visual themes.
    \item Image translation -- turning semantic maps into images (e.g. GauGAN).
    \item Semantic segmentation for turning images into semantic masks (e.g. U-Net).
    \item Unsupervised semantic segmentation for extracting semantic features (e.g. Tile2Vec).
    \item Texture synthesis -- creating large patterns based on a smaller sample (e.g. InGAN).
\end{itemize}
\end{abstract}

\section{Introduction}

Advances in deep learning \cite{goodfellow_deep_2016} offer new ways to create and modify visual content. They can be used for entertainment, including video game development. This paper presents a survey of various machine learning techniques that game developers can directly use for texture and level creation and improvement. These techniques offer a way to:
\begin{itemize}
    \item Generate infinite content that closely resembles real data.
    \item Change or modify the visual style.
    \item Save the time for artists level designers and artists by extending the board or filling features.
\end{itemize}

These techniques differ from a classical approach of procedural content generation \cite{shaker_procedural_2016} using hand-written rules, which is the keystone of some genres (e.g. roguelike games). In contrast, machine learning models learn from the training dataset \cite{summerville_procedural_2018,liu_deep_2021}.

While we focus on images, it is worth mentioning progress in text-generation neural networks (e.g. transformers such as GPT-2 \cite{radford_language_2019} and GPT-3 \cite{brown_language_2020}). They are powerful enough not only to supplement entertainment, e.g. with writing poetry \cite{branwen_gpt-2_2019} but be the core of a text adventure game such as AI Dungeon \cite{raley_playing_2020,walton_ai_2019}. In this game, all narration is AI-generated and allows any text input by the user (rather than a few options). GPT-3 network is powerful enough to generate abstract data (such as computer code) and may be able to create spatial maps encoded as text (e.g. as XML or JSON). One of its extensions, DALL·E \cite{ramesh_zero-shot_2021}, is a text-image model able to generate seemingly arbitrary images from the description, such as \emph{“an armchair in the shape of an avocado”}.

Recent progress in reinforcement learning resulted in creating networks surpassing top human players in Go \cite{silver_general_2018}, reaching grandmaster level in real-time strategy StarCraft II \cite{vinyals_grandmaster_2019}, and playing at a decent level in a multiplayer battle arena Dota~2 \cite{openai_dota_2019}. These methods learn from interacting with the game environment, learning human-like strategies \cite{jaderberg_human-level_2019}, and exploiting unintended properties of game physics and level maps \cite{baker_emergent_2020}. Other notable approaches include automatic level evaluation \cite{davoodi_approach_2020} and level generation from gameplay videos \cite{guzdial_toward_2016}. For some networks, level generation is an auxiliary task to improve an AI model performance \cite{risi_increasing_2020}.
Generating levels can be modified by parameters to adjust game difficulty or address particular needs in educational games \cite{wardaszko_mobile_2017}, offering the potential to increase their effectiveness \cite{park_generating_2019}.

We present seven approaches to create level maps, each using statistical methods, machine learning, or deep learning. In particular, we include:
\begin{itemize}
    \item Generative Adversarial Networks for creating new images from existing examples.
    \item Super-resolution techniques for upscaling images while preserving crisp detail.
    \item Neural style transfer for changing visual themes.
    \item Image translation -- turning semantic maps into images.
    \item Semantic segmentation for turning images into semantic masks.
    \item Unsupervised semantic segmentation for extracting semantic features.
    \item Texture synthesis -- creating large patterns based on a smaller sample.
\end{itemize}

In this paper, we put a special emphasis on methods that are readily available for use. The most popular deep learning frameworks, including TensorFlow \cite{abadi_tensorflow_2016} and PyTorch \cite{paszke_pytorch_2019}, are open-source. Moreover, it is common that the newest models are shared as GitHub code, or even directly as Google Colaboratory notebooks, allowing users to use them online, without any setup. Other models can be accessed via backend, as a service, or with TensorFlow.js \cite{smilkov_tensorflowjs_2019} using GPU support in the browser.

\section{Choice of methods}

Numerous methods are (or can be) used for game development. We focused on ones, which fulfill the following set of criteria:

\begin{itemize}
    \item Learn from data rather than use hand-written rules.
    \item Work on or with images (that is, any 2D spatial data).
    \item Can be used for generating, enhancing, or transforming levels.
    \item Can directly fit into the workflow of level designers and artists \cite{guzdial_friend_2019}.
    \item Are state of the art, in the sense that there is no other method that obviously outclasses them.
\end{itemize}

We choose to restrict to a set of methods that can be related to each other in a meaningful way and focus on the new statistical methods.  While we consult state-of-the-art surveys \cite{stojnic_browse_nodate}, we are aware that there are no straightforward, numeric benchmarks for content generation. What matters is the desired effect by game developers and a subjective experience by the players.

Text generation, procedural or not, is vital for a wide range of adventure or text-based games. Comparing such methods deserves another study, both due to the sheer volume of the topic and the difficulty of relating the text to images (as it is \emph{"apples to oranges"}).

\begin{figure}
    \centering
    \includegraphics[width=0.6\textwidth]{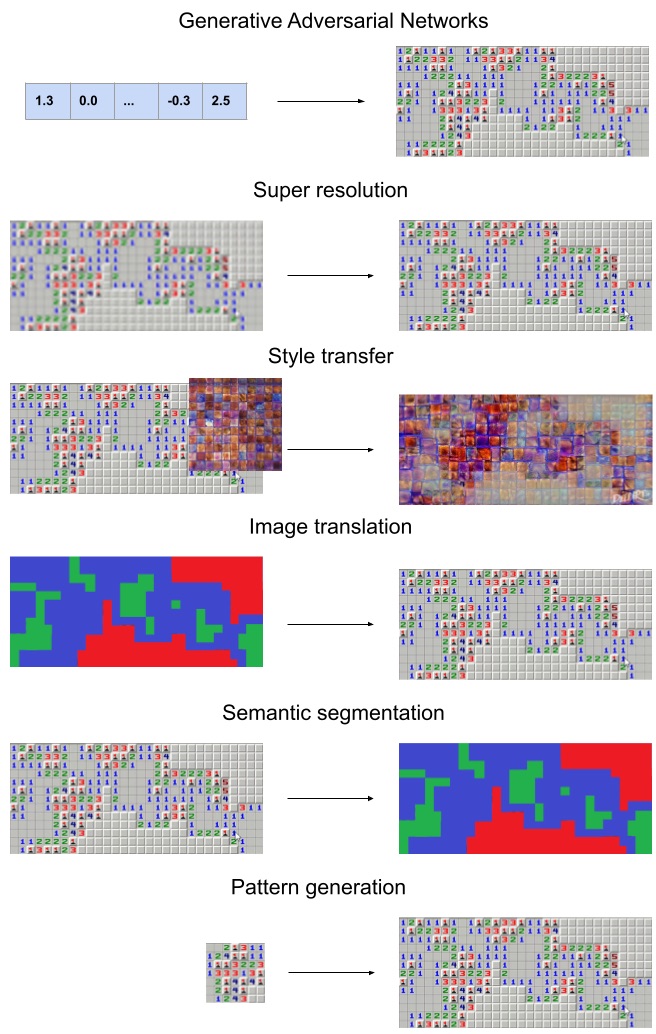}
    \caption{A conceptual summary of transformations performed by the discussed methods. We show a list of manually created input-output pairs.}
    \label{fig:methods}
\end{figure}


We acknowledge that practically any statistical, machine learning or deep learning method can have a potential for game development. However, we aim to cover ones for which their use is straightforward. Since tools used in commercial games are often proprietary secrets, we do not restrict ourselves only to methods with known applications in video games.

It is important to emphasize that these seven method classes are, in general, non-interchangeable. As depicted in \figref{fig:methods}, each requires a different input, produces a different output, and requires a different setup (which usually involves a specific training dataset). They solve different problems. Hence, our motivation is not to compare the quality of the results but to provide use-cases when a particular method has an edge. Even though we use the term “image”, we are not restricted to 2D images with 3 color channels. The same methods work for more channels (e.g. RGBA, RGB + heightmap, or feature maps representing other physical properties) and 3D voxel data.

\section{Description of methods}

\subsection{Generative Adversarial Networks}

Regular Generative Adversarial Networks (GANs) generate new images similar to ones from the training set \cite{goodfellow_generative_2014,goodfellow_nips_2017}. The core idea is that there are two networks -- one generating images (Generator) and the other distinguishing generated images from real ones (Discriminator) \cite{kahng_gan_2019}. However, their training is notoriously difficult and requires much effort to create any satisfactory results. GANs take as an input a vector, which might be random, fixed, or generated from the text \cite{reed_learning_2016,zhang_stackgan_2017,dash_tac-gan_2017}. 

\begin{figure}
    \centering
    \begin{tabular}{cc}
        \includegraphics[width=0.6\textwidth]{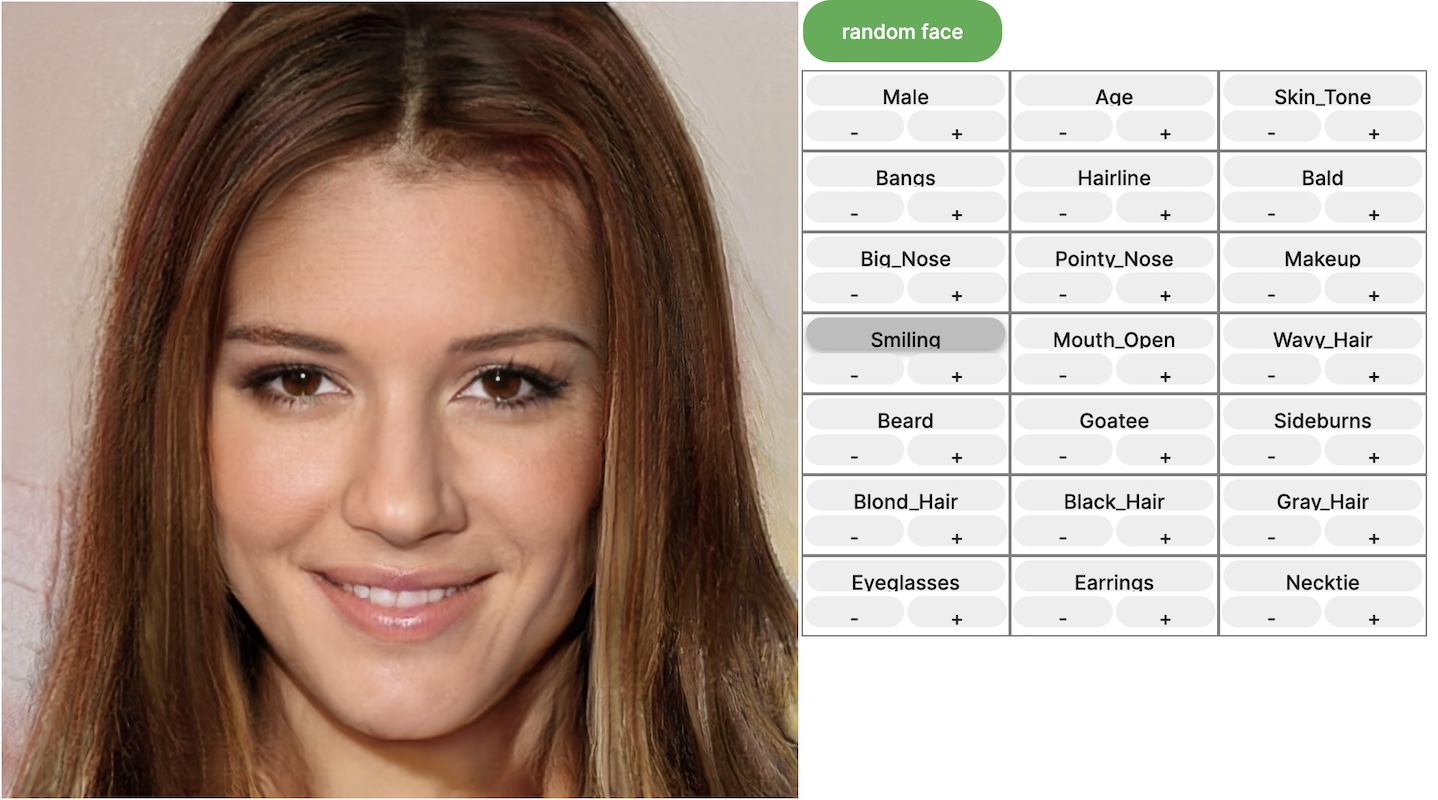} &
        \includegraphics[width=0.28\textwidth]{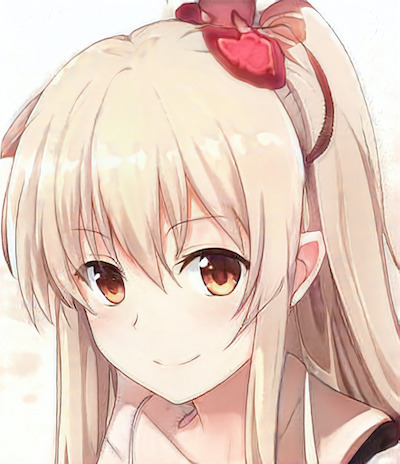}
    \end{tabular}
    
    \caption{A photorealistic face generated from a set of human-interpretable parameters \cite{guan_summitkwantransparent_latent_gan_2018} (left), and a similarly created anime face \cite{branwen_this_2019,branwen_this_2019-1} (right).}
    \label{fig:progan}
\end{figure}

Historically, networks were restricted to generating small images (e.g. $64\times64$ or $128\times128$ pixels) \cite{jin_towards_2017,brundage_malicious_2018,brock_large_2018}.
Only recent progress allowed to generate large-scale (e.g. $1024\times1024$) detailed images such as human faces with ProGAN \cite{karras_progressive_2018}, as seen in \figref{fig:progan}. Its training process involves a series of neural network training stages. It starts from high-level features (such as $4\times4$ images showing an extremely blurred face), step-by-step learning to include high detail. This training is crucial, as otherwise convolutional neural networks tend to overemphasize local features (visible by humans or not) and neglect high-level features.

This class offers potential for one-step creation of levels, from the high-level overview to low-level features. These are extremely data and computing-intensive. However, there is an ongoing effort to make them more data-efficient \cite{karras_training_2020}. Moreover, most examples are based on fairly standardized datasets, e.g. faces of actors or anime characters \cite{branwen_making_2019} -- with consistent style and color. The same approach might not work for more diverse collections of images or photos.

\subsection{Super-resolution (image upscaling)}

Video games often benefit from high visual detail, unless the low resolution is a conscious design choice -- as for some retro-styled games.
Super-resolution \cite{sun_learned_2020,anwar_deep_2020} offers a way to improve texture details by a factor (typically 2x or 4x) without additional effort. Unlike more traditional approaches (nearest neighbor, linear, bicubic), it preserves crisp detail. It is not possible to flawlessly generate more data. However, it is possible to create plausible results –- crisp details resembling authentic, high-resolution images. The method can be applied in two ways: upgrading textures or improving the game resolution in real-time.

\begin{figure}
    \centering
    \begin{tabular}{cc}
        Original & Upscaled x4  \\
        \includegraphics[width=0.3\textwidth]{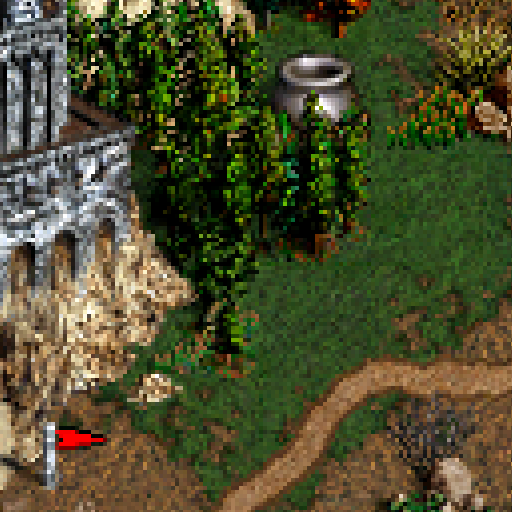} &
        \includegraphics[width=0.3\textwidth]{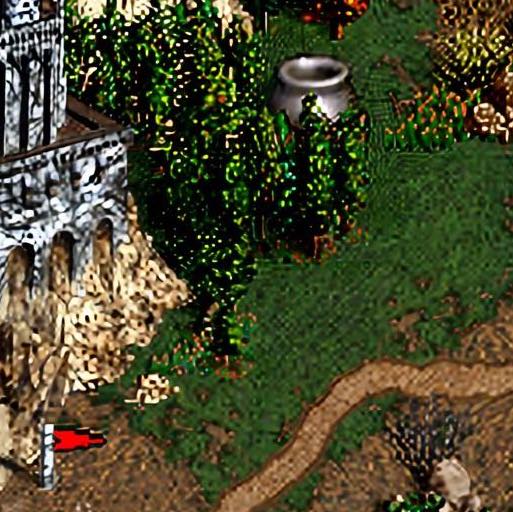}
    \end{tabular}
    \caption{Upscaling with ESRGAN a scene from Heroes of Might and Magic III.}
    \label{fig:esrgan}
\end{figure}

\begin{figure}
    \centering
    \includegraphics[width=0.9\textwidth]{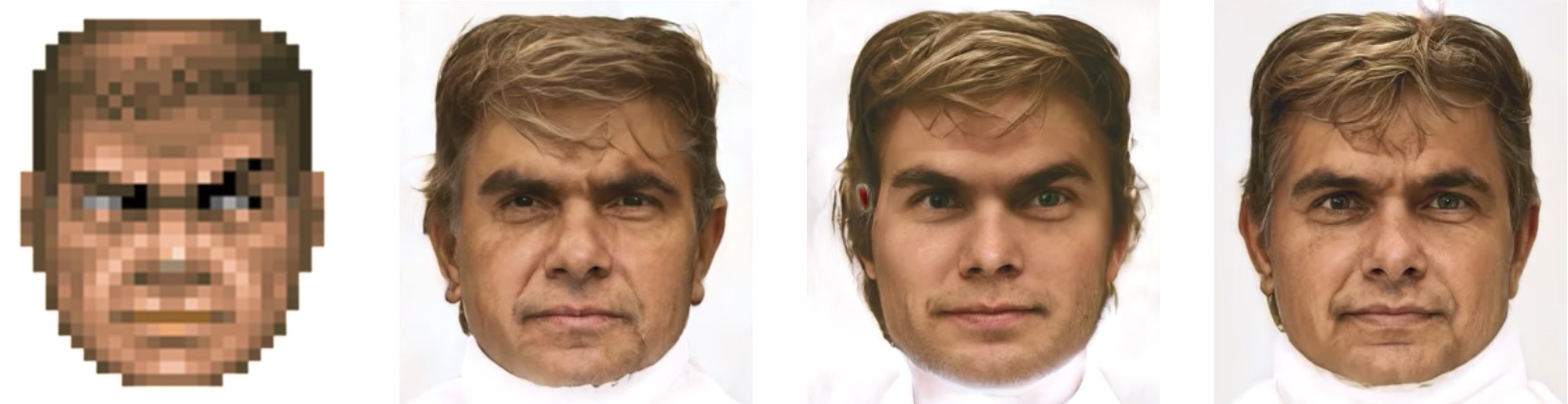}
    \caption{Depixelizing Doomguy with PULSE \cite{bycloud_depixelizing_2020}, reproduced with permission. Note that there are many plausible faces (right) consistent with the original, low-resolution version from Doom (1993) (left).}
    \label{fig:pulse_doomguy}
\end{figure}

A super-resolution network ESRGAN \cite{wang_esrgan_2019} became widely used as it was released with pre-trained weights and easy-to-use code. It can improve older games and save artists’ time for new games as only a lower resolution would be required. Fans are already doing it – both as a proof-of-principle (e.g.for a single screenshot as in \figref{fig:esrgan}) and to create playable mods that enhance game graphics. There are channels dedicated to upscaling games \cite{noauthor_ai_2018,noauthor_gameupscale_2018} that resulted in fan-made textures usable in games such as the 3D shooter Return to Castle Wolfenstein \cite{hart_krischan74rtcwhq_2019}. Compare with a seemingly manual approach of remastering old games, e.g. Diablo II: Resurrected \cite{mccaffrey_diablo_2021} and StarCraft: Remastered \cite{hafer_starcraft_2017}. In addition to generic upscaling networks, the is a PULSE \cite{menon_pulse_2020} network that offers upscaling of faces up to x64 times as in \figref{fig:pulse_doomguy}.

Another key application is improving resolution in real-time. Given that we have fixed computing power (usually restricted by the GPU), it improves the trade-off between resolution and frames per second. That is, we can improve resolution while keeping the same frame rate or improving the frame rate while keeping the same resolution. Algorithms such as DLSS 2.0 by Nvidia \cite{burnes_nvidia_2020,burgess_rtx_2020} are already used by AAA games such as Cyberpunk 2077 \cite{castle_here_2021}.

\subsection{Style transfer}

\begin{figure}
    \centering
    \begin{tabular}{cccc}
        & 
        \includegraphics[width=0.23\textwidth]{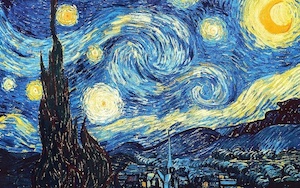} & 
        \includegraphics[width=0.23\textwidth]{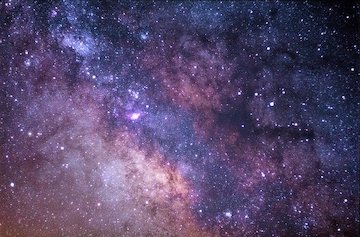} & 
        \includegraphics[width=0.17\textwidth]{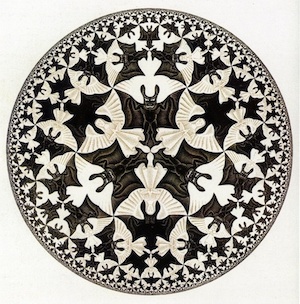}
        \\
        \includegraphics[width=0.23\textwidth]{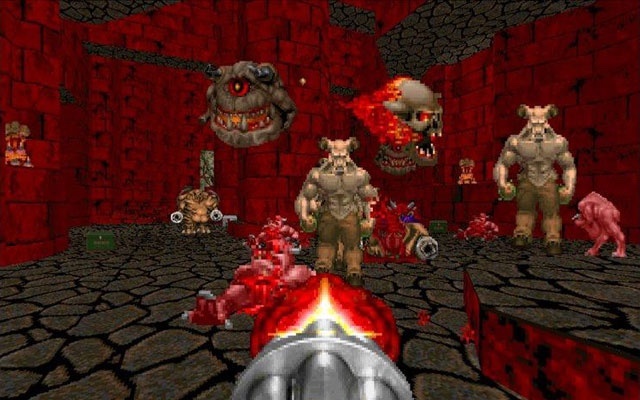} &
        \includegraphics[width=0.23\textwidth]{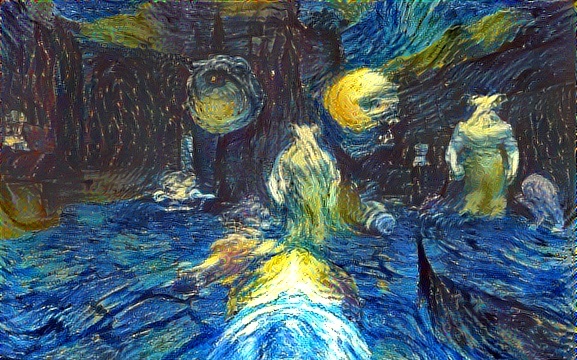} & 
        \includegraphics[width=0.23\textwidth]{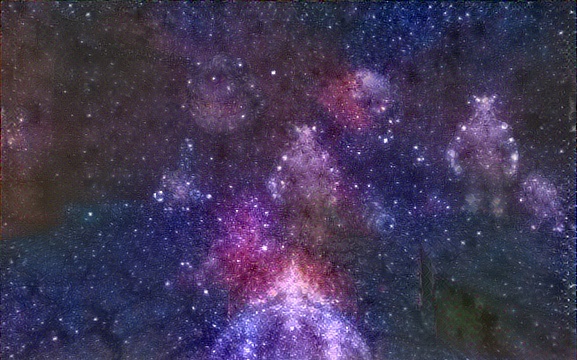} & 
        \includegraphics[width=0.23\textwidth]{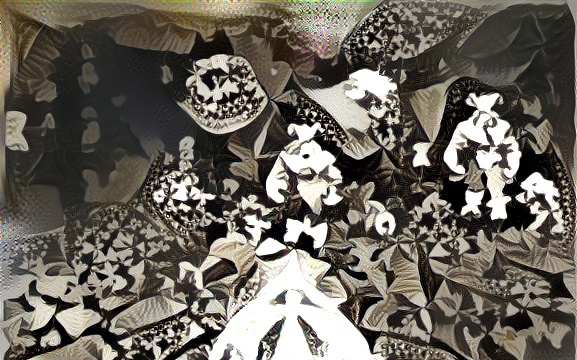}
    \end{tabular}
    
    \caption{An original screenshot from Doom (1993) and three styles: \emph{The Starry Night} by van Gogh, a constellation photo by Pexels, and \emph{Angels and Devils} by Escher (top for the style, bottom for the modified image). Generated online with DeepArt \cite{bethge_deepartio_2015}.}
    \label{fig:style_doom}
\end{figure}

Instead of increasing resolution, we may want to change the style of the image. For example, to turn foliage into dunes or daytime textures into nighttime. The simplest example of style transfer involves taking two images (one with content and the other with style) and generating a painting-like picture \cite{gatys_neural_2016} or video \cite{ruder_artistic_2016-1}. These methods work best for creating dreamy or trippy visual styles but can go well beyond that \cite{semmo_neural_2017,jing_neural_2020}. It is already accessible as a plugin for a popular game engine, Unity3D \cite{deliot_real-time_2020}. Style transfer can be used for post-processing, as depicted in \figref{fig:style_doom}, or as part of an asset generation workflow \cite{allen_neural_2016}.

Newer approaches (usually involving GANs) can change crucial features without sacrificing quality \cite{luan_deep_2017,an_ultrafast_2020}. They can be sensitive and detailed features as a gradual change of faces (e.g. by age or ethnicity) while maintaining photorealistic quality \cite{karras_style-based_2019,karras_analyzing_2020}.

These techniques look promising for changing textures or adding a real-time filter. It includes turning rendered graphics into realistic ones, reassembling real photos, e.g. with GTA V as an example \cite{richter_enhancing_2021}. While such transformations improve the realism of some pieces (e.g. foliage), they may reduce the saturation, cleanliness, and other desirable features of game graphics.

\subsection{Image translation}

Image translation is a broad subject of using neural networks to turn one image into another. Technically, a few other methods can be viewed as such: style transfer, super-resolution, or feature extraction. This section focuses on turning feature maps into levels, fixing feature maps, enriching maps with details (trees), or adding additional modalities (e.g. depth) from existing fields.

\begin{figure}
    \centering
    \begin{tabular}{cc}
        \includegraphics[width=0.3\textwidth]{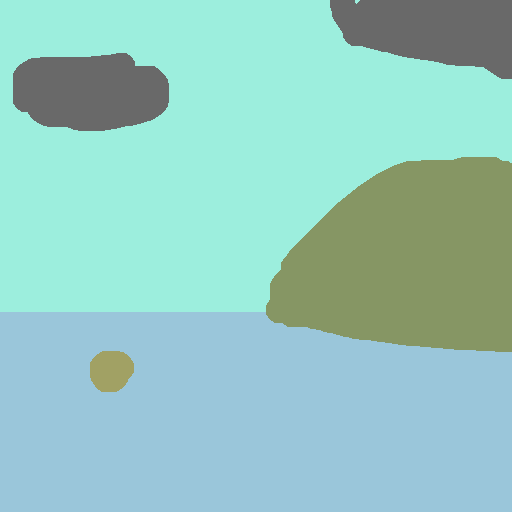} &
        \includegraphics[width=0.3\textwidth]{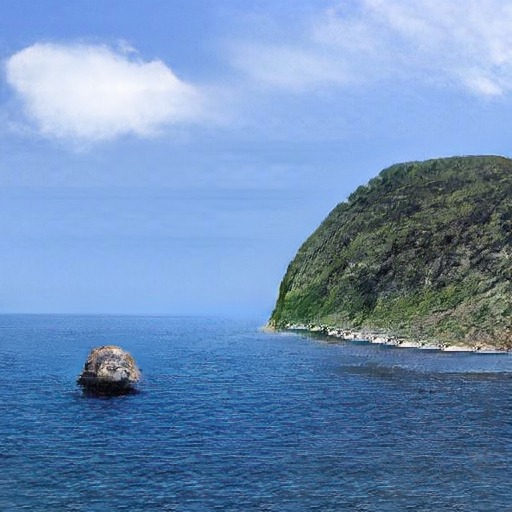}
    \end{tabular}
    
    \caption{Nvidia GauGAN beta \cite{park_semantic_2019} for turning paintbrush-like feature drawings into realistic images. Colors corresponding: sea, sky, rock, clouds and mountains (left) are turned into a photorealistic picture (right).}
    \label{fig:gaugan}
\end{figure}

Notable examples include pix2pix \cite{isola_image--image_2017} and CycleGAN \cite{zhu_unpaired_2017}, translating between various classes of images. GauGAN \cite{park_semantic_2019} is a network focused on turning semantic maps into realistic images -- see \figref{fig:gaugan}. Another similar application is turning 2D graphics into a pseudo-3D perspective -- even without any data obtained from a game engine \cite{shih_3d_2020}. Moreover, abstract game graphics can be turned into photorealistic landscapes -- e.g. for Minecraft \cite{hao_gancraft_2021}.

\subsection{Supervised image segmentation}

\begin{figure}
    \centering
    \begin{tabular}{cc}
      \includegraphics[width=0.3\textwidth]{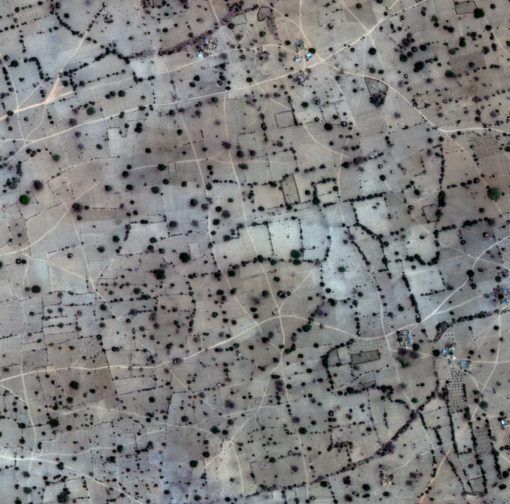} &
      \includegraphics[width=0.3\textwidth]{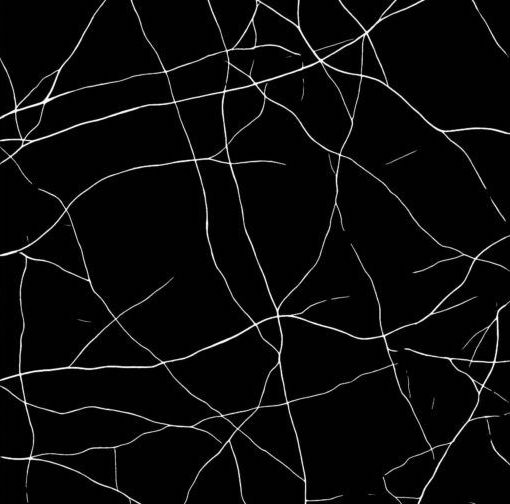}
    \end{tabular}
    \caption{Supervised image segmentation with a custom U-Net: turning satellite images (left) into road maps (right) \cite{nowaczynski_deep_2017}, reproduced with permission.}
    \label{fig:unet_deepsense}
\end{figure}

Turning real photos into game maps usually requires manual work. Semantic segmentation is a class of models extracting semantic features from images \cite{ronneberger_u-net_2015,chen_deeplab_2018}. For example, to change an aerial image into areas with different properties as depicted in \figref{fig:unet_deepsense}. In a race car game, it would be extracting the location of roads, gravel, grass, foliage, and water. A semantic map can be applied to change the game behavior (e.g. different friction on each zone) or generate a map with different textures or graphics, but with overall structure taken from a real-world example.

In supervised semantic segmentation, we need to have training data in the form of pairs: the input image and semantic features, with the same resolution, with one or more classes. Network architectures such as U-Net typically need a relatively small training dataset (as for each image, they treat every pixel as a data point) and a simple training process.

\subsection{Unsupervised image segmentation}

\begin{figure}
    \centering
    \includegraphics[width=0.8\textwidth]{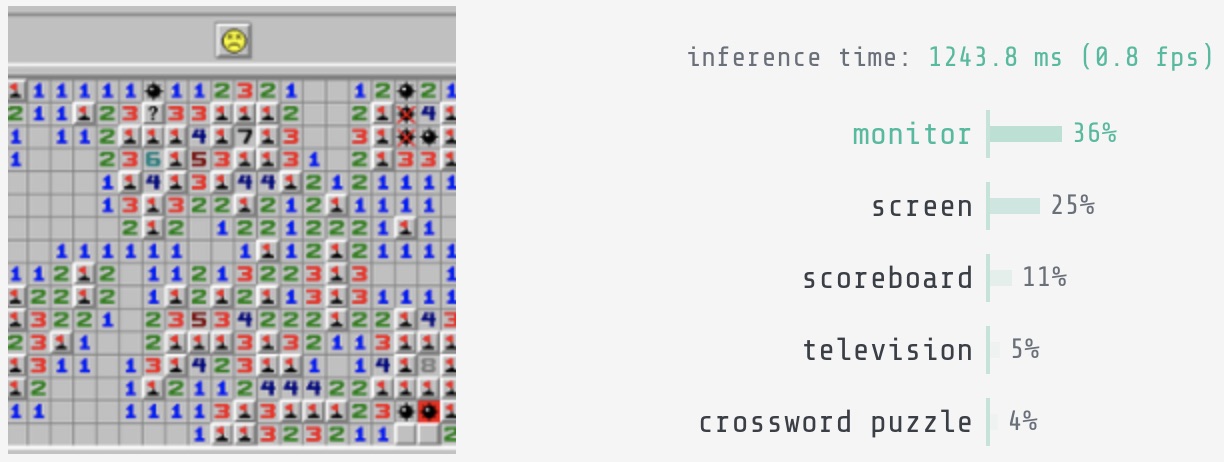}
    \caption{An example that ResNet-50 trained on ImageNet dataset does not work well with Minesweeper, using online Keras.js \cite{chen_kerasjs_2017}.}
    \label{fig:resnet50_minesweeper}
\end{figure}

Manual labeling of all data with semantic maps can be labor-intensive and rigid. For a typical project, one would need to label a considerable fraction of data, which makes it infeasible if we want to lower the workload of level designers.
Unsupervised feature extraction can be created in a few different ways. The most common approach is transfer learning \cite{bengio_deep_2011} -- that is, using visual patterns extracted by pre-trained networks \cite{olah_feature_2017} such as ResNet-50 \cite{he_deep_2016}. However, in many cases, game graphics differ substantially from real photos in style, scale, and content -- see \figref{fig:resnet50_minesweeper}. Given that convolutional neural networks are very sensitive to low-level patterns, any digital artistic style may mislead them. Moreover, abstract graphics (e.g. Minesweeper) or of a different scale (e.g. SimCity) may be non-suitable for networks primarily trained on ImageNet \cite{kornblith_better_2019} -- a dataset of photos taken with regular cameras.

In this case, we may need to resort to self-supervised and unsupervised methods trained on our dataset. For that, we may use autoencoders \cite{masci_stacked_2011}, unsupervised clustering \cite{caron_deep_2018}, or methods extracting content based solely on visual pattern proximity such as Tile2Vec \cite{jean_tile2vec_2019} and Patch2Vec \cite{fried_patch2vec_2017}.

\begin{figure}
    \centering
    \begin{tabular}{cc}
        \includegraphics[width=0.4\textwidth]{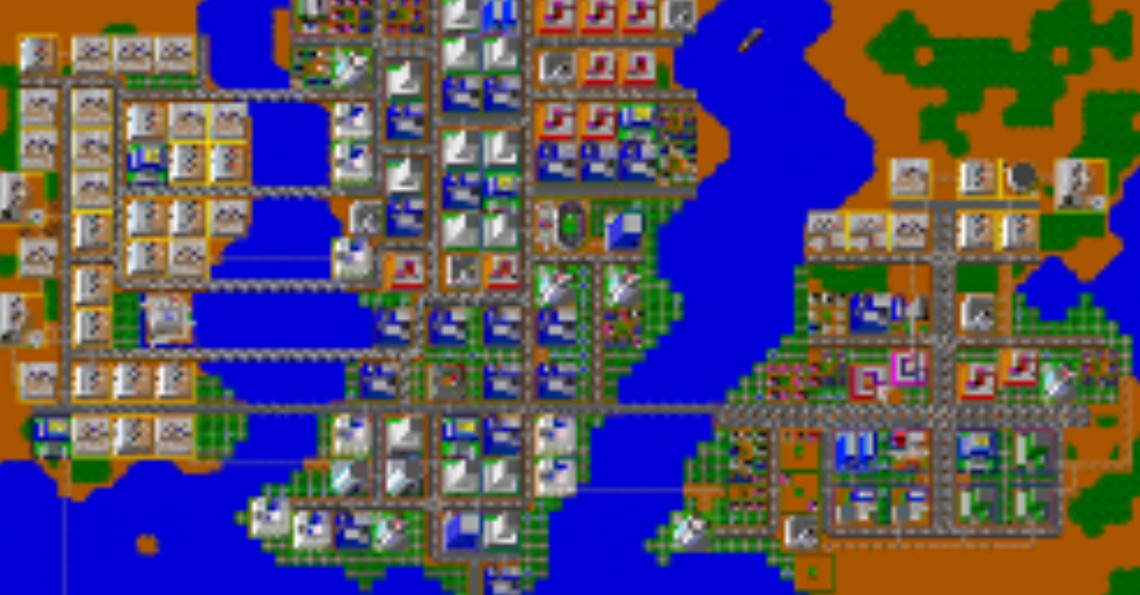} &
        \includegraphics[width=0.4\textwidth]{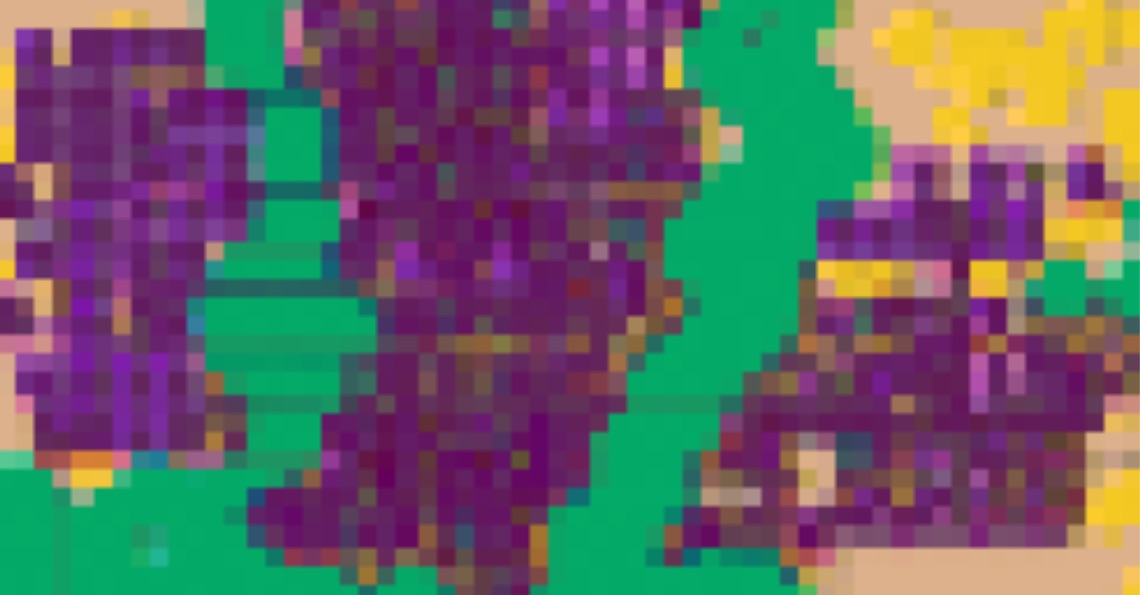} \\
        \includegraphics[width=0.4\textwidth]{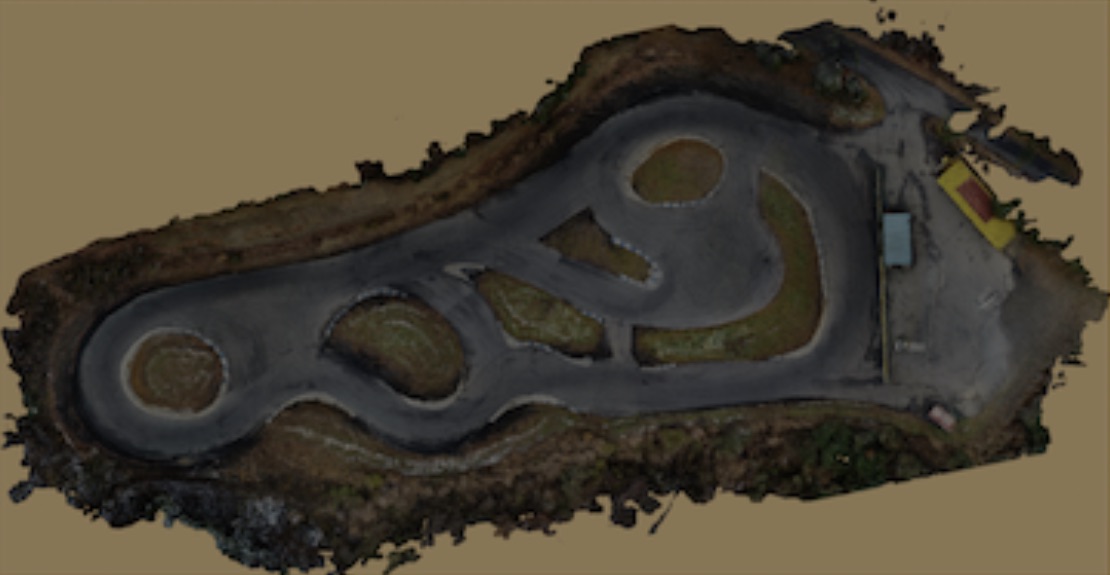} &
        \includegraphics[width=0.4\textwidth]{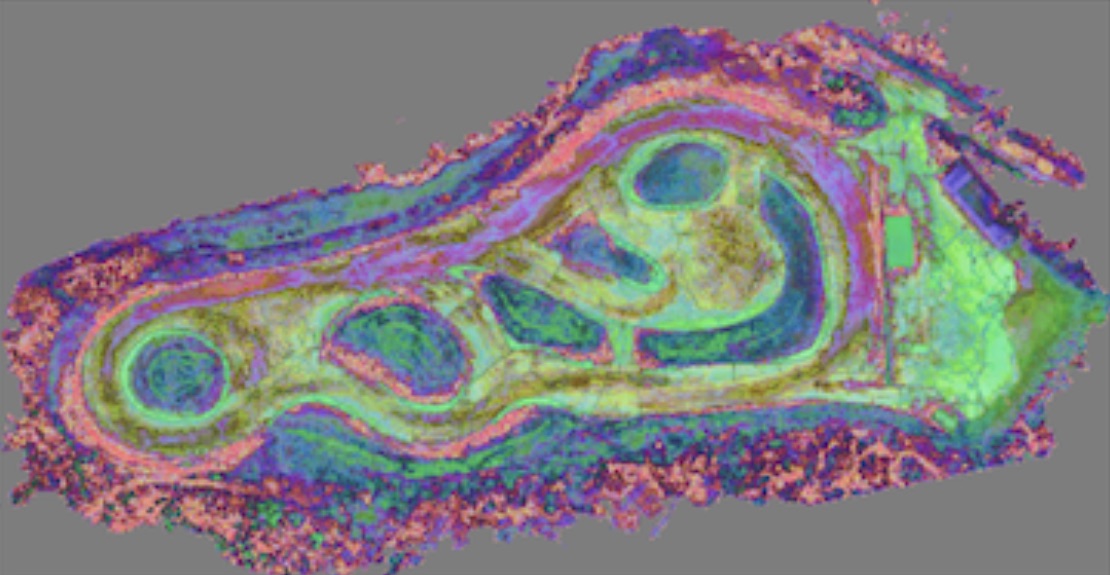}
    \end{tabular}
    \caption{One-shot results for a SimCity map and an aerial scan of a racetrack -- input images (left) and three largest Principal Components mapped to RGB channels (right).}
    \label{fig:eccgames}
\end{figure}

At ECC Games SA, we use a similar technique \cite{migdal_unsupervised_2020} for turning aerial photos of race track by drone into semantic maps that allow both to distinguish robust features (e.g. road vs foliage) and details (density and angle of tire trails). This technique can work even on a single image, with no pre-training, which offers extreme data efficiency rare in deep learning. In particular, it means that there is no extra effort on image labeling or generating large datasets. We show two diverse examples in \figref{fig:eccgames}.

\subsection{Pattern synthesis}

Pattern synthesis is a method used to expand a spatial pattern up to an arbitrary size. Conceptually, it relates to super-resolution, but it makes the image larger instead of going deeper in detail.
A straightforward approach involves generating bigger textures without self-repetition to give them a natural feel \cite{zhou_non-stationary_2018,shocher_ingan_2019}. This technique can be applied statically or for creating patterns in real-time \cite{liang_real-time_2001}. Another way to go is to work at the level of the general map -- i.e. expanding high-level semantic maps from a smaller sample. Except for deep learning, some classical algorithms are widely used, such as WaveFunctionCollapse \cite{gumin_mxgmnwavefunctioncollapse_2016,kim_automatic_2019,karth_wavefunctioncollapse_2017}.

\begin{figure}
    \centering
    \begin{tabular}{cc}
        \includegraphics[width=0.2\textwidth]{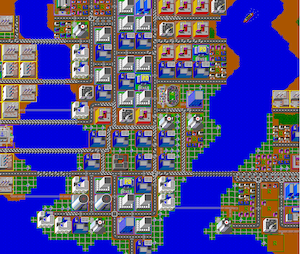} &
        \includegraphics[width=0.333\textwidth]{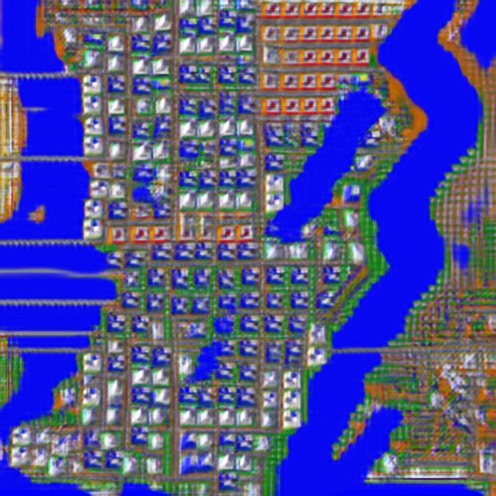}
    \end{tabular}
    \caption{Pattern synthesis using InGAN \cite{shocher_ingan_2019} based on a single map from SimCity. The map is treated purely as an image. Splitting into semantic regions (e.g. tiles related to buildings) would improve the result.}
    \label{fig:synth}
\end{figure}

For game genres such as classic platformer games and scrolling shooters, 2D level maps can be considered a sequence of 1D stripes. We can tackle generating 1D sequences with additional, simpler methods. We can tackle generating 1D sequences with additional, simpler methods, e.g. the classical statistical model of Hidden Markov Chain \cite{rabiner_tutorial_1989}. Deep learning models for sequence generation, such as a Long Short-Term Memory network (LSTM) \cite{hochreiter_long_1997} and a Gated Recurrent Unit (GRU) \cite{cho_learning_2014}, offer more flexibility and even better results, especially for longer sequences. The 1D patterns of images can be exploited by transformers generating images strip-by-strip \cite{chen_generative_2020}. For example, they can generate Super Mario levels generated with an LSTM \cite{summerville_super_2016}.

\section{Conclusion}

We presented an array of deep learning methods for the creation and enhancement of levels and textures. They show the potential to improve the game development process, offer new experiences to the players, or generate content in real-time. As we have seen in the past, technical advancements allow not only to improve existing game formulas, but also to create new genres \cite{kushner_masters_2003}.

Depending on the problem we tackle or an enhancement we want to make -- we need to consider different types of deep learning models.

\begin{enumerate}
    \item We need to translate the task into an input-output problem. Both inputs and outputs can be photos or other real-world images, in-game graphics, semantic maps, or any other format of spatial encoding of low-level textures and high-level maps.
    \item We need to check classes of models and if they provide the transformation we need. In this paper, we discussed seven different classes, which provide a good starting point.
    \item We need to see if any model within this family provides results of the desired quality. This step may involve a simple plug \& play setup with an existing trained model or involve a much more complicated process of finding a specific model and training it with our data.
\end{enumerate}

Quite a few deep learning techniques are already used in game development -- by fan communities, modders, professional game developers, or as a tool provided by game engines.
While progress in deep learning is fast, even current models offer broad perspectives for supporting game development -- with many applications and opportunities yet to be discovered. The abundance of publications from the last two years (and not-yet-published novel results) shows that it is a fruitful and quickly developing field. The latter resorts us to citing GitHub repositories and other sources to reference the newest methods.


\section*{Acknowledgments}

We would like to thank the ECC Games team, with special thanks to Marcin Prusarczyk and Piotr Wątrucki for the project overview, and Bartłomiej "Valery" Walendziak for sharing his insights on the stages of level development from the artist’s perspective. We are grateful to Olle Bergman, Anna Olchowik and Jan Marucha for their comments and to  Rafał Jakubanis for his extensive editorial feedback on the draft.

We focus on creating race track maps from aerial scans of real race tracks. This task involves a few levels of complexity -- from the high-level general pattern of the map to low-level details and  decorations such as tire tracks. 
This research is supported by the EU R\&D grant POIR.01.02.00-00-0052/19 for project  \emph{“GearShift -- building the engine of the behavior of wheeled motor vehicles and map generation based on artificial intelligence algorithms implemented on the Unreal Engine platform -- GAMEINN”}.




{\small
\bibliographystyle{habbrv_unsorted}
\bibliography{zotero}
}

\end{document}